%% file: main.tex
\documentclass[runningheads]{llncs}

\usepackage{eccv}

\usepackage{eccvabbrv}
\usepackage[misc]{ifsym}
\usepackage{graphicx}
\usepackage{booktabs}
\usepackage{multirow}
\usepackage[accsupp]{axessibility}  
\usepackage{amsmath}
\usepackage{hyperref}
\usepackage{orcidlink}

\begin{document}

\title{Rethinking Reward Signals in Video GRPO: When Scores Become Targets} 
\author{
Rui Li\inst{1,*} \and
Yuanzhi Liang\inst{2,*} \and
Ziqi Ni\inst{3} \and
Haibin Huang\inst{2} \and
Chi Zhang\inst{2} \and
Xuelong Li\inst{2(\mbox{\Letter})}
}
\authorrunning{R. Li et al.}

\institute{
University of Science and Technology of China, Hefei, China\\
\email{rui.li@mail.ustc.edu.cn}
\and
Institute of Artificial Intelligence (TeleAI), China Telecom, Shanghai, China
\and
Southeast University, Nanjing, China
\\
\email{liangyzh18@outlook.com, xuelong\_li@ieee.org}\\
\textsuperscript{*} Equal contribution. \quad
\textsuperscript{\Letter} Corresponding author.
}





\maketitle

\begin{abstract}
Group Relative Policy Optimization (GRPO) enables stable and preference-oriented updates via group-wise comparisons for post-training video generation. However, GRPO directly optimizes reward-induced advantages. Under sustained optimization, the reward score can lose fidelity as a proxy for true video quality, consistent with the phenomenon described by Goodhart’s Law. This leads to two recurring issues: (i) shortcut-driven optimization under composite objectives and (ii) reward saturation within prompt groups. To address these issues, we introduce TaRoS, a Target-Robust Reward Signaling framework for Video generation GRPO. 
TaRoS leverages component level performance assessment together with intra-group sparsity to organize multi-aspect rewards towards optimization objectives. In addition, it adaptively downweights components that exhibit saturation, thereby preserving effective optimization directions and mitigating redundancy.
This maintains meaningful optimization directions and preserves within‑group ranking separation, thereby preventing reward hacking and leading to more reliable policy updates. Extensive experiments show consistent improvements in visual fidelity, motion coherence, and text-video alignment over strong baselines. 
\end{abstract}

\input{sections/1_intro}
\input{sections/2_related}

\input{sections/3_method}
\input{sections/4_experiment}

\input{sections/5_disscussion}

\input{sections/6_conclusion}


\bibliographystyle{splncs04}
\bibliography{main}
\end{document}

%% file: sections/1_intro.tex
\section{Introduction}
Group Relative Policy Optimization (GRPO) is particularly suitable for video post-training. It samples multiple candidates per prompt, performs group-wise comparisons, and updates the generator using group-relative advantages. This design makes GRPO inherently reward-centric. The reward is not only used to evaluate outputs but also determines the relative advantages that drive the policy gradient. As a result, the reward functions as an optimization target rather than a passive critic, and GRPO’s behavior is largely determined by the reward signal~\cite{mroueh2025reinforcement,wu2025rewarddance}.
\begin{figure}[htbp]
    \centering
    \includegraphics[width=1\linewidth]{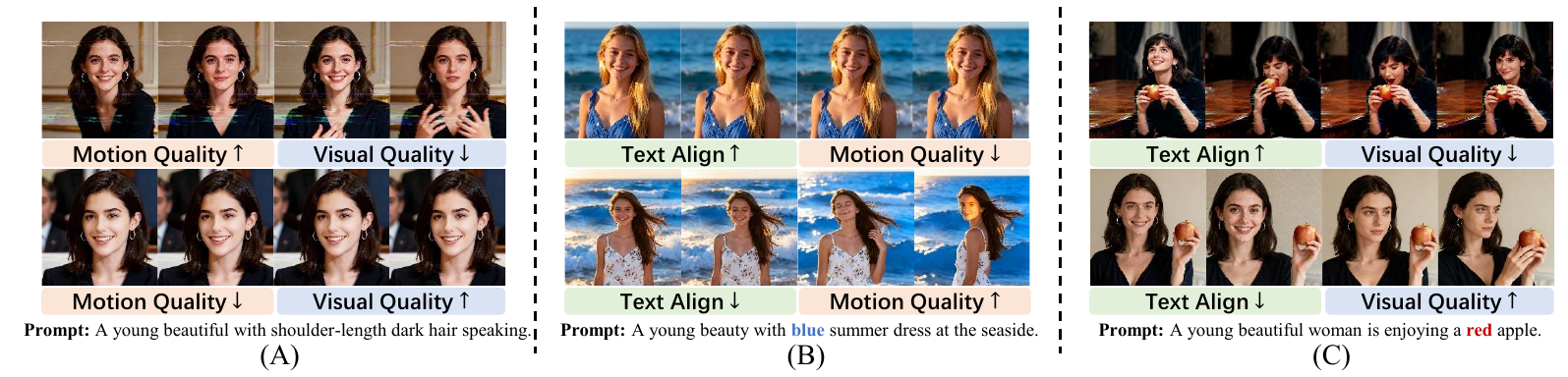}
\caption{Dilemma (i): Goodhart's Law under video generation GRPO perspective with composite reward. Case A: visual quality can be sacrificed to boost motion. Case B: optimizing text alignment reduces motion, while improving motion degrades text alignment. Case C: optimizing text alignment may degrade textual fidelity.}
    \label{fig:dilemma1}
\end{figure}
This places video generation GRPO in a regime aligned with Goodhart’s law: when a measure becomes a target, optimization pressure can weaken its correlation with the intended objective. In GRPO, this is a structural problem because reward scores are directly used to form advantages and drive updates. As a result, the failure is not only a matter of using a “better” reward model or a “better” optimizer in isolation. It can emerge whenever the evaluator is repeatedly optimized against in a closed loop. Notably, such breakdowns can appear even when the reward behaves reasonably under static, offline evaluation~\cite{karwowski2023goodhart,christiano2017deep}.

This paper rethinks video generation GRPO from a simple question: what does it take for a reward signal to remain useful once it is placed inside the optimization loop? We argue that the core issue is not only how to build a stronger reward model. The key is whether the reward remains a reliable training signal throughout the optimization trajectory. In GRPO, learning is driven by within-prompt, group-wise reward differences. These differences determine (i) which candidates are preferentially reinforced relative to others, thereby defining the optimization direction, and (ii) the magnitude of the update, which corresponds to the effective advantage signal. As a result, reward signaling in GRPO typically breaks in two corresponding ways: it can misdirect optimization under composite objectives, or it can lose separation as score differences compress over training.

With this framing, we identify two recurring failure modes in video GRPO:

(1) Directional failure under composite objectives. Video quality is multi-aspect, including visual fidelity, motion coherence, etc. A common strategy is to scalarize these heterogeneous criteria via a fixed weighted reward. Under GRPO, this scalarization often induces shortcut ascent directions. The model can increase one component while degrading or neglecting others\cite{zhang2025r1,yang2024rewards,xu2024visionreward,liu2025improving}. For example, as shown in Fig.~\ref{fig:dilemma1}, when motion-related rewards are optimized too early or too strongly, the policy may prefer high-frequency flicker or unstable camera shake that increases “motion” scores but reduces visual quality and consistency\cite{huang2024vbench}. Similarly, alignment-oriented rewards can be improved via superficial cue matching while the visual content remains implausible or low-fidelity.

\begin{figure}[htbp]
    \centering
\includegraphics[width=1\linewidth]{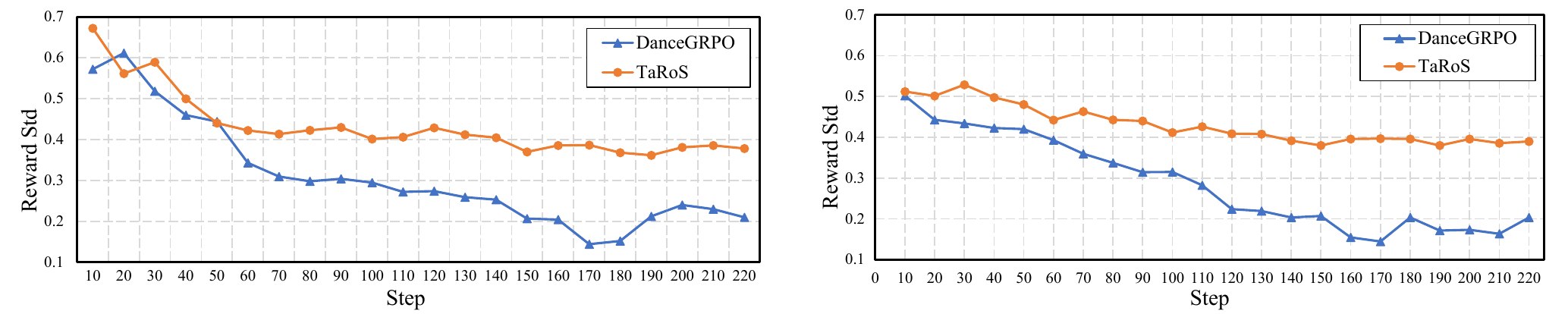}
\caption{Dilemma (ii): Reward Saturation. Fixed reward weights tend to cause early reward saturation. (Left) Qwen-2.5-VL-72B; (Right) VideoAlign. TaRoS maintains higher reward variance than the baseline throughout training, indicating its ability to alleviate saturated components and significantly delay reward saturation.}
\label{fig:dilemma2}
\end{figure} 
(2) Discriminability failure along the training trajectory. GRPO relies on within-group ranking. It needs the reward to separate sampled candidates for the same prompt. As illustrated in Fig.~\ref{fig:dilemma2}, reward scores in practice tend to compress as training progresses, leading to frequent ties and diminishing margins between samples. When the reward cannot reliably distinguish candidates, group-relative advantages become low signal-to-noise. Updates then become uninformative or even noise-driven. For example, once a reward saturates on a prompt family, multiple candidates receive nearly identical scores. GRPO then lacks a stable preference gradient and may drift due to small perturbations\cite{wang2026tagrpo}.

This motivates a practical definition of a “useful” reward signal in GRPO. A reward is useful only if it provides (i) a meaningful optimization direction and (ii) sufficient within-group separation to support stable advantages. If either property fails, GRPO updates become unreliable. This also explains why reward selection and reward scheduling are coupled. A reward that looks strong offline may become uninformative or misdirecting once it is optimized against. Conversely, scheduling alone cannot fix a judge that is systematically biased or lacks resolution on the relevant training distribution.

Based on these observations, we develop TaRoS, a Target-Robust Reward Signaling framework for video GRPO to address both failure modes. TaRoS models the reward scale and reorganizes multi-aspect reward signals to adaptively adjust the overall optimization direction, thereby mitigating shortcut-driven optimization. Additionally, TaRoS monitors whether each reward component remains informative for group-wise comparison, ensuring that saturated signals do not dominate updates. This monitoring is achieved by leveraging both performance assessment of reward components and the sparsity of intra-group signals, which together provide a dynamic calibration mechanism for robust reward signaling. Evaluator properties further influence both optimization direction and score separation through systematic bias, sensitivity, and coverage. We therefore adopt a generalist vision–language evaluator as the judging backbone, which provides more stable rankings across prompts and training stages and supports consistent reward signaling across components.

Our contributions are threefold. (1) We provide a Goodhart-driven diagnosis of GRPO reward failure in video generation, isolating two recurring mechanisms: shortcut-dominated optimization under composite objectives and loss of within-group ranking resolution during training. (2) We propose TaRoS, a Target-Robust Reward Signaling framework that structures multi-aspect reward signals and adaptively suppresses non-informative components, improving robustness of GRPO updates. (3) Extensive experiments show stronger stability and consistent improvements in visual quality, motion coherence, and text–video alignment over competitive baselines.

%% file: sections/2_related.tex
\section{Related Work}
\noindent\textbf{Post Training of Visual Synthesis.} 
Early efforts in optimizing visual synthesis primarily relied on Proximal Policy Optimization (PPO)~\cite{furuta2024improving}. These methods first trained a value network to capture specific human preferences, then used the trained reward model to score the generated videos, and performed preference fine-tuning based on these scores. However, PPO-based approaches require maintaining a value network, which leads to low training efficiency. Building upon this idea, Group Relative Policy Optimization~\cite{shao2024deepseekmath,dancegrpo,liu2025flow} extends preference optimization to multi-modal generative tasks. By leveraging group-based relative comparisons rather than single-sample rewards, GRPO can capture richer preference signals and achieve more robust alignment in visual synthesis models.

\noindent\textbf{Multi-Stage Approaches.}  
Multi-stage optimization has been widely adopted to progressively refine generative models across spatial, temporal, and semantic dimensions. For example, Spatial-then-temporal~\cite{li2023spatial} first enhance spatial and appearance features using static images, and then incorporate temporal information through video reconstruction with a distillation loss to preserve spatial representations. TTC~\cite{liu2025future} improve spatial quality via masked region restoration and subsequently enhance temporal consistency with a three-frame sampling strategy and an auxiliary branch. Enhance-A-Video~\cite{luo2025enhance} further enhance temporal consistency by emphasizing off-diagonal elements in the temporal attention map and scaling cross-frame attention values. These pre-training studies highlight the potential of multi-stage design in post-training for video generation. 
However, existing approaches listed above rely on rigid, manually designed optimization stages, and such handcrafted stage scheduling often lacks generalizability\cite{huang2026exposurebiasalleviatedirectional}. 

\noindent\textbf{Reward Models.} 
Recent studies~\cite{hu2023reinforcement,furuta2024improving} have introduced IQA models and VLMs as reward functions, enabling semantically grounded reinforcement learning for visual generation. For instance, VILA~\cite{lin2024vila} has been applied to construct preference pairs for human alignment~\cite{yang2025ipo}, while VideoAlign~\cite{liu2025improving} is employed to enhance both visual quality and motion quality in text-to-video generation~\cite{dancegrpo}. Similarly, InFLVG~\cite{fang2025inflvg} leverages frame-wise consistency and CLIP-based similarity to facilitate coherent long-form video generation. Despite these advances, existing approaches typically rely on fixed and single-dimensional reward signals, which are prone to reward hacking. To mitigate this, VRG~\cite{prabhudesai2024video} and InstructVideo\cite{yuan2024instructvideo} leverage diverse reward models (e.g., CLIP~\cite{radford2021learning}, HPSv2~\cite{wu2023human}, and other vision-language or video representation models~\cite{tong2022videomae,schuhmann2022laion,fang2021you,assran2025v}) to jointly optimize semantic alignment, perceptual quality, and temporal coherence.
Nevertheless, as discussed above, jointly employing multiple reward models inevitably introduces conflicts, and balancing the supervision strength across heterogeneous rewards remains a challenging problem.
Approaches such as \cite{mercier2018stochastic,yu2020gradient} rely on complex gradient manipulations, which incur substantial memory overhead. In the context of large-scale video generation models under distributed training, such methods are nearly infeasible.

Building on these observations, we highlight the need for a reward signal design that adapts over the course of training, while resists early saturation.
~\label{sec:related}

%% file: sections/3_method.tex
\begin{figure*}[tp]
    \centering
    \includegraphics[width=1\linewidth]{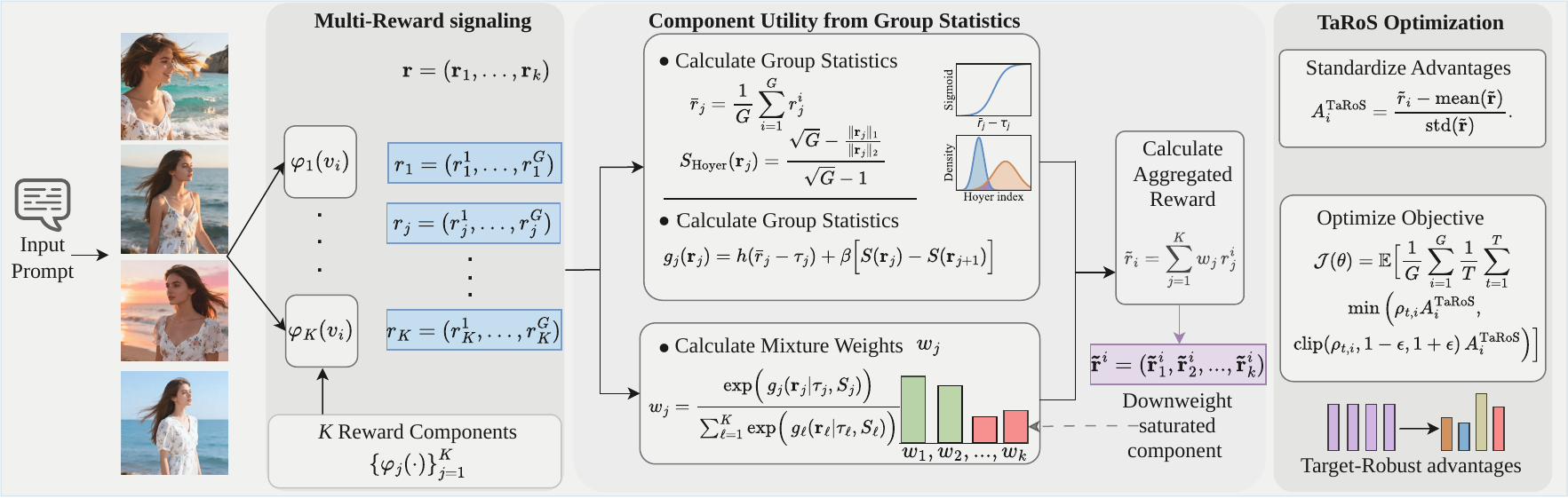}
    \caption{Pipeline of the proposed TaRoS. Our method adaptively reorganizes multi-aspect rewards. For each sample, TaRoS tailors the policy gradient direction. Rewards are aggregated and gated through target-robust reward signaling. The normalized reward signals are then used to compute advantages that guide gradient-based fine-tuning of the generator.}
\label{pipeline}
\end{figure*}

\section{Preliminary}
\noindent\textbf{SDE Sampling.} In ODE sampling, the sampling process is deterministic, formulated as
\begin{equation}
    dx_t = v_t \, dt,
\end{equation}
where $v_t$ denotes the control variable at time $t$, without any stochastic component.

However, since reinforcement learning requires stochasticity to encourage exploration, such a deterministic model is not sufficiently flexible for exploration in GRPO. 
Unlike ODE sampling, which follows a single deterministic trajectory, SDE sampling introduces stochastic perturbations that encourage exploration and prevent collapse to suboptimal modes. It incorporates both a drift term and a diffusion term.
\begin{equation}
    dx_t = \underbrace{\left( v_t(x_t) - \tfrac{\sigma^2}{2} \nabla \log p_t(x_t) \right)}_{\text{drift term}} dt 
    + \underbrace{\sigma_t \, dw}_{\text{diffusion term}},
\end{equation}
where the \emph{drift term} governs the deterministic dynamics of the process, while the \emph{diffusion term} introduces stochastic perturbations through Brownian motion, thereby enhancing exploration.



\section{Target-Robust Reward Singnal framework (TaRoS)}
GRPO updates are driven by group-relative advantages computed from within-prompt reward differences. Thus, effective training requires reward signals that both guide optimization in a useful direction and preserve within-group separation. In practice, scalarized composite rewards can encourage shortcut directions, while reward scores often compress over training and reduce ranking resolution. TaRoS constructs a target-robust aggregated reward by weighting multiple reward components using their within-group informativeness. This suppresses saturated components and yields more reliable GRPO updates.

\subsection{Target-Robust Reward Signaling}
For each prompt, GRPO samples a group of $G$ videos $\{v_i\}_{i=1}^{G}$, where $v_i=\{f_t^i\}_{t=1}^{T}$.
We define $K$ reward components $\{\varphi_j(\cdot)\}_{j=1}^{K}$, each evaluating a distinct aspect of generation quality.
For component $j$, the reward for candidate $i$ is
\begin{equation}
r_j^i = \varphi_j(v_i), \qquad \mathbf r_j = (r_j^1,\ldots,r_j^G).
\end{equation}

\subsection{Component Utility from Group Statistics}
GRPO updates depend on within-group reward differences. We therefore score each reward component by how informative it is for group-wise comparison.

For component $j$, we compute the group mean
\begin{equation}
\bar r_j=\frac{1}{G}\sum_{i=1}^{G} r_j^i ,
\end{equation}
and measure within-group separation using the Hoyer sparsity index~\cite{hoyer2004non}:
\begin{equation}
S_{\mathrm{Hoyer}}(\mathbf r_j)=
\frac{\sqrt{G}-\frac{\|\mathbf r_j\|_1}{\|\mathbf r_j\|_2}}
{\sqrt{G}-1}, \qquad S_{\mathrm{Hoyer}}(\mathbf r_j)\in[0,1].
\end{equation}
Low values indicate score compression within the group, implying weak ranking resolution for GRPO.

We define the utility of component $j$ from group statistics as
\begin{equation}
g_j(\mathbf r_j)
= h(\bar r_j-\tau_j)
+ \beta\Big[S_{\mathrm{Hoyer}}(\mathbf r_j)-S_{\mathrm{Hoyer}}(\mathbf r_{j+1})\Big],
\end{equation}
where $h(\cdot)$ is a smooth activation function (e.g., sigmoid), $\tau_j$ is a component-specific threshold, and $\beta$ controls the contribution of the separation term. For the last component $j=K$, we set $S_{\mathrm{Hoyer}}(\mathbf r_{K+1})=0$, so the separation term reduces to $\beta\,S_{\mathrm{Hoyer}}(\mathbf r_K)$. 
The first term $h(\bar r_j-\tau_j)$ serves as a component level performance assessment that measures whether the component has reached a meaningful quality level in the current group, while the second term encourages components that retain stronger within-group ranking resolution than the subsequent (typically higher-level) component.
Overall, $g_j$ prioritizes reward components that are both effective and informative for GRPO-style group comparison.
This design ties reward integration to what GRPO fundamentally depends on: stable and well-separated within-group reward differences that yield informative group-relative advantages.

\subsection{Soft Integration of Reward Components}
TaRoS converts utilities into mixture weights via a soft selection rule:
\begin{equation}
w_j=
\frac{\exp\!\big(\, g_j(\mathbf r_j)\big)}
{\sum_{\ell=1}^{K}\exp\!\big(\, g_\ell(\mathbf r_\ell)\big)},
\qquad \sum_{j=1}^{K} w_j = 1,\quad w_j\in[0,1],
\end{equation}
The aggregated reward for candidate $i$ is then
\begin{equation}
\tilde r_i = \sum_{j=1}^{K} w_j\, r_j^i.
\end{equation}
This continuous weighting avoids hard stage boundaries and reduces the influence of components whose group-wise scores have saturated and provide limited separation.

\subsection{Optimization with TaRoS Reward}
Let $\tilde{\mathbf r}=(\tilde r_1,\ldots,\tilde r_G)$ denote the aggregated rewards within a group.
We standardize them to form TaRoS advantages:
\begin{equation}
A_i^{\mathrm{TaRoS}}=
\frac{\tilde r_i-\operatorname{mean}(\tilde{\mathbf r})}
{\operatorname{std}(\tilde{\mathbf r})}.
\end{equation}
Standardization improves comparability across prompt groups and prevents scale differences from dominating optimization.

We then optimize a GRPO-style clipped objective using $A_i^{\mathrm{TaRoS}}$:
\begin{equation}
\begin{split}
\mathcal{J}(\theta) =
\mathbb{E}\Big[
\frac{1}{G}\sum_{i=1}^{G}\frac{1}{T}\sum_{t=1}^{T}
\min\Big(
\rho_{t,i} A_i^{\mathrm{TaRoS}},
\mathrm{clip}(\rho_{t,i},1-\epsilon,1+\epsilon)\, A_i^{\mathrm{TaRoS}}
\Big)
\Big],
\end{split}
\end{equation}

where the importance ratio is
\begin{equation}
\rho_{t,i}=
\frac{\pi_\theta(a_{t,i}\mid s_{t,i})}
{\pi_{\theta_{\mathrm{old}}}(a_{t,i}\mid s_{t,i})}.
\end{equation}

Additionally, TaRoS is agnostic to the specific form of reward components $\{\varphi_j\}$.
Following common practice in recent GRPO-based video post-training, we instantiate several components using a vision--language model (VLM) as the reward evaluator.
We provide comparisons across different evaluator backbones and component choices in Sec.~Experiment.

%% file: sections/4_experiment.tex
\section{Experiment}

\begin{table*}[t]
\centering
\caption{Quantitative VBench results for Wan2.1‑T2V‑1.3B, we compare Wan and DanceGRPO with our \textbf{TaRoS}. 
The best score for each metric is shown in \textbf{bold}.}
\label{wan1.3B_vbench_result}
\resizebox{\textwidth}{!}{
\setlength{\tabcolsep}{4pt}
\begin{tabular}{lcccccc}
\toprule
\multirow{2}{*}{\textbf{Metric}} 
& \multirow{2}{*}{\textbf{Wan1.3B}} 
& \textbf{DanceGRPO} &\textbf{DanceGRPO}
& \textbf{Ours} & \textbf{Ours} & \textbf{Ours}
 \\
\cmidrule(lr){3-7} \cmidrule(lr){7-7}
& &\textbf{Videoalign} & \textbf{Qwen2.5VL-72B} & \textbf{Videoalign}& \textbf{Qwen2.5VL-7B} & \textbf{Qwen2.5VL-72B}  \\
\midrule

    Aesthetic quality &60.92&58.99 &59.79 & 60.88 & 61.87 & \textbf{62.64}\\ 
    Appearance style &20.41 &20.70&20.60 & 20.82&\textbf{21.45} & 21.26 \\ 
    Human action &74.00  &73.00 &75.00 &74.00 &\textbf{76.00} &\textbf{76.00}    \\ 
    Image quality &67.65 &66.61 &67.31 & 67.92 &67.83 &\textbf{68.49}  \\ 

    Multiple objects &59.14  &57.54 &59.16 &59.73 &60.46  &\textbf{63.49}    \\  
    Object class &74.84&77.68 &77.32 & 76.42&\textbf{77.93} &74.21    \\  
    Overall consistency  &23.63 &23.68 &23.61 &23.67 &\textbf{23.76}  &23.67 \\  
    Scene &22.38 & 25.79& 25.22 & 24.62& 20.49 &\textbf{25.94}   \\  
    Spatial relationship &68.08 &62.89 &66.34 & 67.16&72.78  &\textbf{72.00}    \\ 

    \addlinespace
    Quality score &83.15 &82.81 &82.83 &83.23 &83.03  &\textbf{83.66}  \\  
    Semantic score &65.31 &66.06 &66.22 &66.26 &67.32  &\textbf{68.15}   \\  
    Total score &79.58 &79.46 &79.72 &79.84 &79.90 &80.56  \\ \bottomrule 
    \end{tabular}
}
\end{table*}

\subsection{Settings} 
\noindent\textbf{Datasets.}
Our training is conducted on the \textit{50k} dataset released by DanceGRPO. For ablation studies, we randomly sampled \textit{1.4k} prompts from the iStock dataset as our test set.

\noindent\textbf{Evaluation Metrics.} 
For quantitative evaluation, we adopt VBench~\cite{huang2024vbench}, which assesses sixteen complementary dimensions to provide a comprehensive understanding of generative model performance. 
For ablation studies, we report four representative metrics, including VideoAlign, where VQ measures visual quality, MQ measures motion quality, and TA measures text alignment. 
In addition, we employ LAION~\cite{schuhmann2022laion} to evaluate video aesthetics. 

\subsection{Implementation details} 
\noindent\textbf{Training Setup.} 
We evaluate TaRoS on Wan2.1-T2V with 1.3B and 14B variants~\cite{wan2025wan}, as well as HunyuanVideo~\cite{kong2024hunyuanvideo}. For each model, the thresholds $\tau$ are calibrated from the historical distribution of group-wise rewards collected during training. TaRoS is evaluator-agnostic and can be instantiated with different VLM-based reward judges. In our experiments, we instantiate reward components using VideoAlign and frozen Qwen2.5-VL~\cite{bai2025qwen2} models (7B and 72B) as alternative evaluator backbones. 

For Wan2.1-T2V-1.3B, videos are sampled at \(480 \times 832 \times 81\) frames at 16 fps, with reward gate thresholds \(\tau_{I} = \tau_{II} = 0.75\), a learning rate of \(5 \times 10^{-6}\), and group size \(G=16\). Training is performed for 220 steps on 16 H100 GPUs.
For Wan2.1-T2V-14B, videos are sampled at \(240 \times 416 \times 53\) frames at 16 fps, with thresholds \(\tau_{I} = 0.75, \tau_{II} = 0.73\), the same learning rate and group size, and training for 220 steps on 32 H100 GPUs. 
For HunyuanVideo-T2V, videos are sampled at \(240 \times 416 \times 53\) frames at 15 fps, with thresholds \(\tau_{I} = 0.70, \tau_{II} = 0.68\), a learning rate of \(2 \times 10^{-6}\), and group size \(G=16\). Training is performed for 100 steps on 32 H100 GPUs. 

\noindent\textbf{Inference.} 
For Wan2.1-T2V-1.3B and Wan2.1-T2V-14B, we configure 50 sampling steps with a resolution of \(480 \times 832 \times 53\) (H\(\times\)W\(\times\)T) at 16 fps. 
For Hunyuan Video-T2V, we adopt 30 sampling steps under the same resolution setting.

\noindent\textbf{Coefficient of Variation.}  
The coefficient of variation is a normalized measure of dispersion, defined as the ratio of the standard deviation to the mean of a dataset. Unlike raw variance or standard deviation, it provides a scale-independent metric, making it suitable for comparing variability across datasets with different units or magnitudes. A lower coefficient of variation indicates that the data points are relatively consistent around the mean, while a higher value suggests greater relative variability. This metric is widely used in statistics and experimental analysis to assess stability, reliability, and the degree of heterogeneity in observations.
\begin{table}[t]
\centering
\caption{Quantitative VBench results for Wan2.1‑T2V‑14B. 
We compare pre‑trained Wan14B baseline with our method. 
The best score for each metric is shown in \textbf{bold}.}
\small
\setlength{\tabcolsep}{4pt}
\scriptsize
\begin{tabular}{lccc}
\toprule
\multirow{2}{*}{\textbf{Metric}} 
& \multirow{2}{*}{\textbf{Wan14B}} 
& \multicolumn{1}{c}{\textbf{Ours}} 
& \multicolumn{1}{c}{\textbf{Ours}} \\   
\cmidrule(lr){3-4}
& & \textbf{Videoalign} & \textbf{Qwen2.5VL‑72B} \\   
\midrule
Aesthetic quality      & 65.33 & \textbf{66.00} & 60.68 \\
Appearance style       & 21.35 & \textbf{21.65} & 21.62 \\
Human action           & 77.00 & 78.00 & \textbf{80.00} \\
Image quality          & 68.03 & 68.06 & \textbf{68.91} \\
Multiple objects       & \textbf{70.27} & 69.05 & 69.81 \\
Object class           & 81.72 & 81.96 & \textbf{82.19} \\
Overall consistency    & 25.08 & 24.97 & \textbf{25.17} \\
Scene                  & 32.12 & \textbf{34.22} & 30.67 \\
Spatial relationship   & 74.97 & 73.79 & \textbf{79.06} \\
\addlinespace
\textbf{Quality score} & 84.03 & 83.61 & \textbf{84.59} \\
\textbf{Semantic score}& 71.18 & 71.94 & \textbf{72.08} \\
\textbf{Total score}   & 81.46 & 81.58 & \textbf{82.09} \\
\bottomrule
\end{tabular}
\label{wan14b_vbench}
\end{table}
\subsection{Quantitative evaluation}
\noindent\textbf{Vbench Results.} 
Table~\ref{wan1.3B_vbench_result} and~\ref{wan14b_vbench} present the VBench evaluation of TaRoS on Wan2.1-T2V. We compare our method with the untrained Wan and DanceGRPO as baselines. 
From the model capacity perspective, larger reward models exhibit fewer semantic biases and later reward saturation, which benefits both the baseline and TaRoS as shown in Table~\ref{wan1.3B_vbench_result}. Nevertheless, even with the 72B reward model, the baseline DanceGRPO achieves only limited improvement. This indicates that scaling reward capacity alone is insufficient without principled reward scheduling. In contrast, TaRoS achieves consistent performance improvements when scaling to larger reward models such as Qwen2.5VL‑7B and 72B, and achieves comparable progress on VideoAlign as well. While DanceGRPO can enhance specific dimensions such as color and background consistency, it often degrades other metrics including image quality and spatial relationships.

We attribute this limitation to the baseline’s direct summation of reward signals during training, which keeps the optimization direction fixed throughout the process. As a result, the generative model can exploit shortcuts to artificially boost evaluation metrics. In contrast, TaRoS dynamically adjusts the weights of reward components, thereby altering the optimization trajectory to some extent and mitigating the risk of reward hacking. More critically, as training proceeds, certain reward components become saturated, and such coarse aggregation injects noise into the reward signal. This noise is particularly detrimental for reinforcement learning with reward feedback under GRPO, as it reduces, or even misjudges, the relative quality of samples within a group. In contrast, our method dynamically adjusts the reward signals according to the quality of sampled candidates, tailoring the optimization stage for each sample and thereby alleviating such optimization dilemmas. Resulting in better performance in Vbench.
\begin{figure}[tp]
    \centering
    \includegraphics[width=1\linewidth]{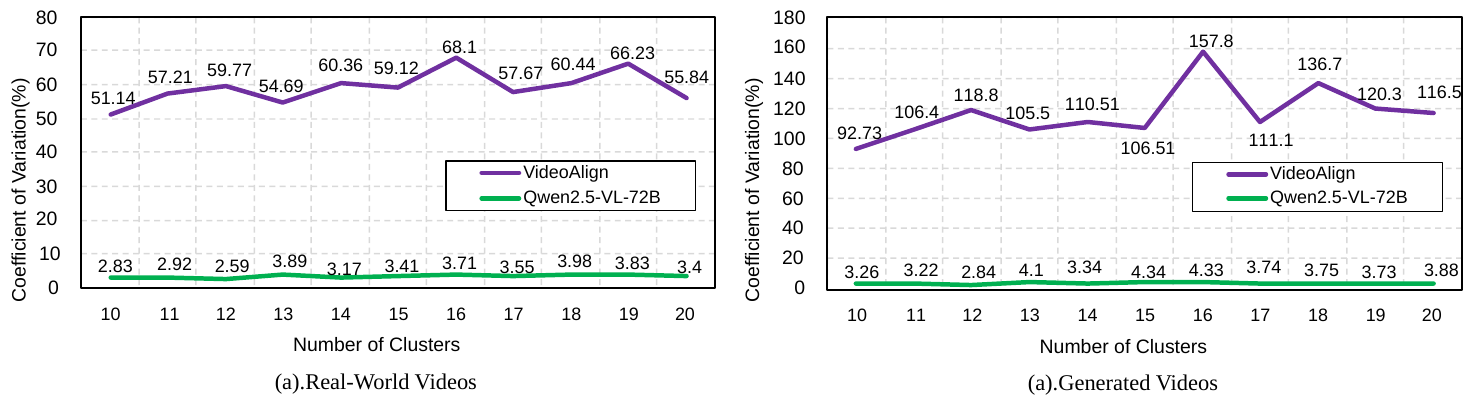}
    \caption{Reward preference analysis. 
We conduct analyses separately on 2k real‑world videos and generated videos, evaluating them with different reward models and reporting the coefficient of variation across semantic categories.}
    \label{fig:videoalgin vs qwen}
\end{figure}

\noindent\textbf{Generalization.}
As shown in Table~\ref{Hunyuan_result}, our method achieves consistent improvements over HunyuanVideo across a broad spectrum of evaluation dimensions. On the visual side, it enhances appearance style, background consistency, and overall image fidelity, yielding generations that are more coherent and aesthetically pleasing. It also strengthens motion smoothness and mitigates temporal flickering, thereby improving dynamic stability and temporal quality. Beyond visual fidelity, our approach demonstrates superior handling of complex scenarios, including multi-object interactions and spatial relationships, while maintaining subject consistency across frames. These results highlight the robustness of our algorithm across diverse backbone architectures. 
\begin{table}[!ht]
    \centering
    \caption{Quantitative results on VBench for HunyuanVideo. 
    We compare HunyuanVideo with our \textbf{TaRoS} using Qwen-2.5VL-72B as reward model. 
    The best performance for each metric is highlighted in \textbf{bold}.}
    \label{Hunyuan_result}
    \resizebox{\textwidth}{!}{
    \begin{tabular}{lccccccccc}
    \toprule
    Model & Appearance & Background & Image & Motion & Multiple & Spatial & Subject & Temporal & Temporal \\
          & style & consistency & quality & smoothness & objects & relationship & consistency & flickering & style \\
    \midrule
    Hunyuan & 19.95 & 96.45 & 64.61 & 98.87 & 69.51 & 67.55 & 95.92 & 99.12 & 24.18 \\
    TaRoS   & \textbf{20.26} & \textbf{96.85} & \textbf{68.83} & \textbf{99.33} & \textbf{71.42} & \textbf{69.84} & \textbf{96.54} & \textbf{99.43} & \textbf{24.24} \\
    \bottomrule
    \end{tabular}}
\end{table}

\begin{figure}[t]
    \centering
    \includegraphics[width=\textwidth]{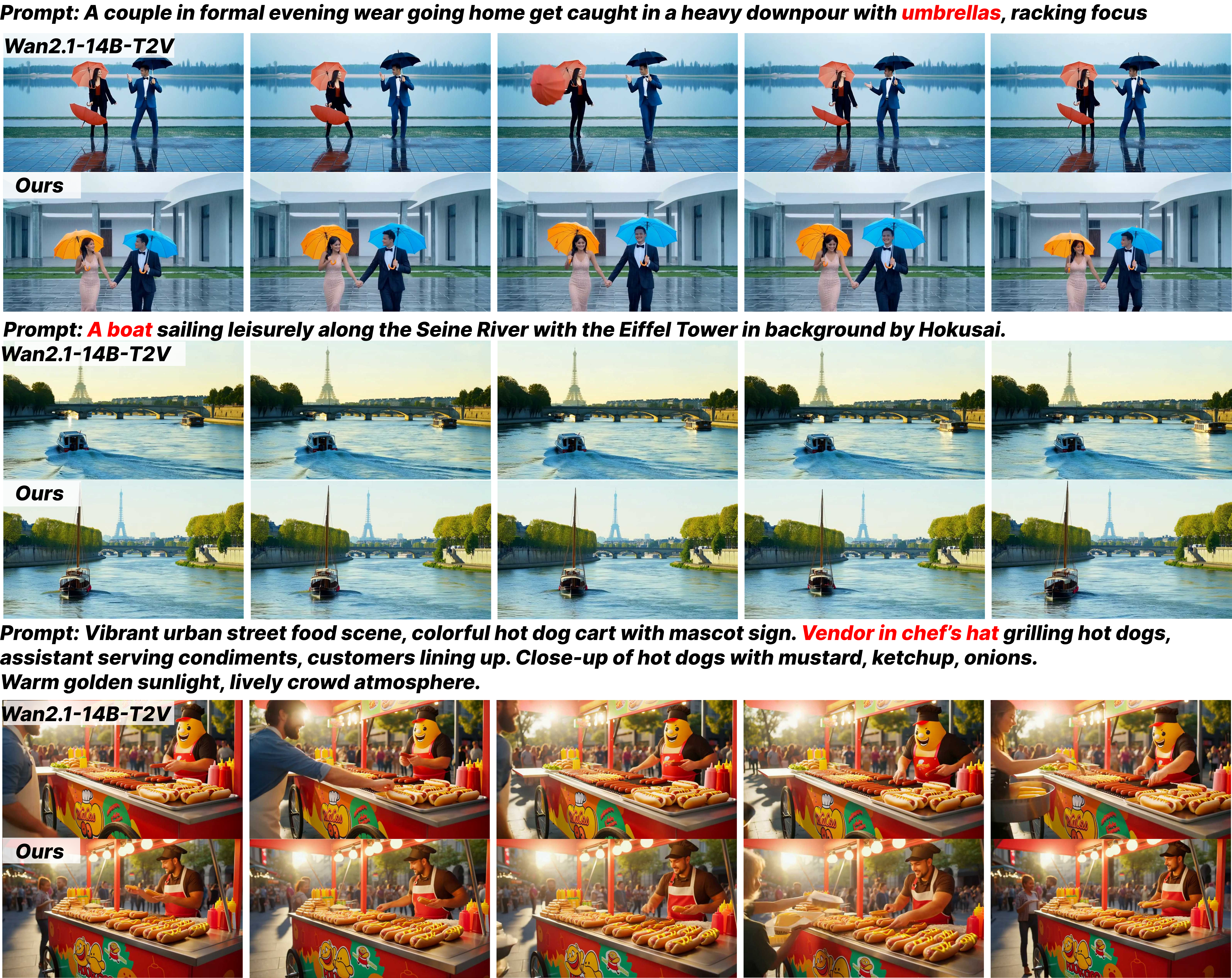}
    \caption{Qualitative comparison between Wan2.1‑T2V‑14B and our fine‑tuned model. 
Top: baseline Wan2.1‑T2V‑14B results. 
Bottom: outputs from our fine‑tuned model.
}
    \label{fig:wan2.1_vs_ours}
\end{figure}

\noindent\textbf{Large Capacity Reward Model.}  
To further justify our choice of larger reward models, we conduct experiments across different model capacities. We randomly sampled \textit{2k} real-world videos with captions from the iStock dataset and generated corresponding videos based on these captions. The videos were grouped into \textit{10–20} semantic clusters using the UMT5~\cite{raffel2020exploring} encoder, and each cluster was scored with both VideoAlign and Qwen2.5VL-72B. We then computed the mean score and the coefficient of variation $\kappa$ within each cluster to quantify evaluation consistency. Rewards were normalized to [0,10] to ensure the validity of the coefficient of variation. The analysis of real-world videos is illustrated in Fig.~\ref{fig:videoalgin vs qwen} (a), where we compare the intra-cluster reward consistency of VideoAlign and Qwen2.5VL-72B across semantic clusters.

Since evaluation is conducted on real-world videos, we expect comparable visual quality across semantic categories. VideoAlign, however, shows large reward variations across clusters, with a coefficient of variation as high as 55.84, revealing bias toward certain content types. In contrast, Qwen2.5VL‑72B yields more consistent scores and a much lower variation of 3.4, indicating stable, content‑invariant evaluations. This bias becomes even stronger on generated videos (Fig.~\ref{fig:videoalgin vs qwen} (b)). Such preference encourages reward hacking during training. For example, models may sacrifice semantic alignment to generate content favored by the reward model. This supports our choice of generalizable, unbiased feedback, as exemplified by Qwen2.5VL‑72B.

\subsection{Qualitative Research}
\noindent\textbf{Visual Comparison.}
To complement the quantitative results, we conduct qualitative case studies that provide intuitive evidence of our method’s advantages. By visually comparing generated samples under different settings, we highlight improvements in visual fidelity, temporal coherence, and semantic alignment—qualities that are not always fully captured by numerical metrics. As illustrated in Fig.~\ref{fig:wan2.1_vs_ours}, case (1) shows that our method prevents incorrect umbrella generation and achieves better image composition. In case (2), it produces higher visual quality with more vivid colors. In case (3), it generates the correct visual theme, demonstrating improved semantic consistency. Overall, these qualitative analyses confirm enhanced visual fidelity and stronger semantic alignment.

\begin{figure}[t]
    \centering
    \includegraphics[width=0.9\linewidth]{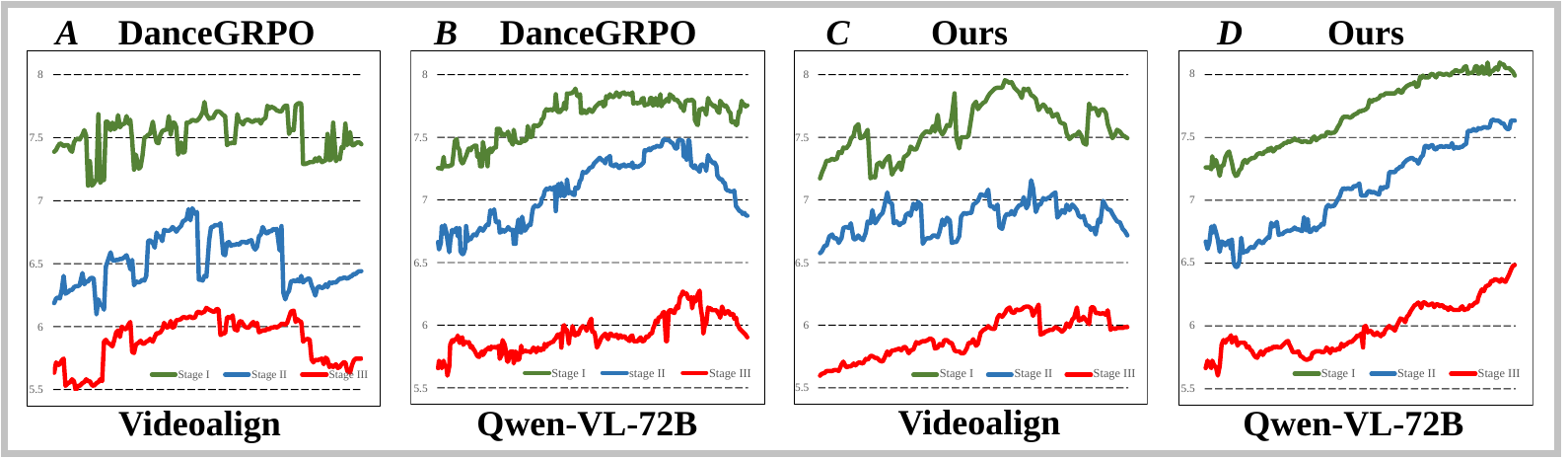}
    \caption{Comparison of reward curves for Wan2.1‑1.3B across various reward models}
    \label{fig:reward cruve}
\end{figure}

\noindent\textbf{Reward Curves.}
To verify the success of our training, we present the learning curves on Wan2.1‑T2V‑1.3B. Following DanceGRPO, all curves are smoothed with a window size of 50 to mitigate extreme outliers. As shown in Fig.~\ref{fig:reward cruve}, the comparison between A and C reveals pronounced reward saturation when VideoAlign is used as the reward model, particularly in the later stages of training. Once reward saturation occurs, the group‑wise advantages that GRPO relies on become unreliable, leading to erroneous gradient updates and eventual training collapse. DanceGRPO suffers from this “collapse‑boost” cycle, in which rewards fluctuate or even oscillate downward, preventing stable convergence. The comparison between A and B further shows that larger reward models can delay saturation, but cannot eliminate it entirely, as saturation inevitably arises with prolonged training. In contrast, the comparison between B and D demonstrates that TaRoS achieves stable reward growth and significantly later saturation. This highlights the superiority of our approach: by suppressing saturated reward components, TaRoS reduces their interference with the overall signal, thereby improving the signal‑to‑noise ratio in reward signal and providing more reliable guidance for gradient updates.

\subsection{Ablation}
\noindent\textbf{(1) Less Dimensions.} 
To investigate whether our algorithm can maintain robustness under reduced reward dimensions, we conducted ablation experiments with fewer reward components. As shown in Table~\ref{ablation}, even with only two dimensions, our method consistently outperforms the baseline, demonstrating its robustness.
Furthermore, in the three-dimensional setting, we observed that danceGRPO improves MQ and TA but degrades VQ, which aligns with our first hypothesis. In contrast, our approach achieves consistent improvements across all four metrics—VQ, MQ, TA (in-domain), and Laion score (out-of-domain). We attribute this to the dynamic adjustment of the optimization gradient direction, which makes it more difficult for the generative model to exploit reward loopholes. Although this does not completely eliminate reward hacking, it effectively mitigates the issue to a significant extent.
\begin{table}[h]
\centering
\caption{Ablation study on less optimization dimension. 
We compare joint training two-stage strategies with DanceGRPO and TaRoS. 
Results are reported on VideoAlign metrics VQ, MQ, and TA, as well as the LAION aesthetic score. }
\scriptsize
\setlength{\tabcolsep}{3pt}   
\renewcommand{\arraystretch}{1} 
\begin{tabular}{lcccc}
\toprule
\textbf{Method} & \textbf{VQ$\uparrow$} & \textbf{MQ$\uparrow$} & \textbf{TA$\uparrow$} & \textbf{LAION$\uparrow$} \\
\midrule
Wan2.1-1.3B-T2V        & 3.448 & 0.2911 & -1.914  & 5.224 \\
DanceGRPO+Two Dimensions         & 3.493 &  0.2973 & -1.409  & 5.233 \\
TaRoS+Two Dimensions             & 3.499 & 0.3166 & -1.312  & 5.261 \\
DanceGRPO+Three Dimensions        & 3.412($\downarrow$) &0.3022($\uparrow$) & -1.172($\uparrow$)  & 5.246 \\
TaRoS+Three Dimensions & 3.501($\uparrow$) & 0.3090($\uparrow$) & -0.7114($\uparrow$) & 5.252 \\
\bottomrule
\end{tabular}
\label{ablation}
\end{table}

\noindent\textbf{(2) Hyperparameter Sensitivity.}
To verify the robustness of the proposed algorithm with respect to hyperparameters $\tau$ and $\beta$, we conducted hyperparameter ablation studies using VideoAlign scoring. As shown in Table~\ref{tab:tau_beta_ablation}, the scores obtained with VideoAlign remain robust under variations of $\tau$ and $\beta$. These results demonstrate that the proposed TaRoS maintains stable performance across different hyperparameter settings, indicating that our method is not overly sensitive to the choice of $\tau$ and $\beta$.

%% file: sections/5_disscussion.tex
\section{Discussion}
\noindent\textbf{Why TaRoS better.}
On the one hand, static weighting fails to provide adaptive scheduling of optimization objectives. Treating all samples with uniform supervision disregards their inherent variability, making the approach both coarse and suboptimal. For example, Enforcing semantic optimation early without a dependable visual foundation creates noisy interference. On the other hand, fixed weights on well-learned prompts cause reward saturation, introducing noise to training progress. TaRoS resolves this by utilizing real-time feedback to tailor the optimization direction. Thus change the optimization direction during training, reducing the risk of reward hacking. Then, TaRoS down-weights saturated reward components, thereby enhancing the signal-to-noise ratio of reward signals during training and delaying reward saturation.

\noindent\textbf{Optimization Order Design.} 
Following prior studies on video quality assessment \cite{sun2025content,ling2025vmbench,rohaly2000video}, we emphasize that visual fidelity is the essential prerequisite for meaningful assessment. Human observers are particularly sensitive to spatial distortions such as blur, noise, or compression artifacts, which dominate subjective judgments of video quality. In contrast, improvements in motion dynamics or semantic alignment can only be reliably perceived when the generated frames already exhibit sufficient perceptual clarity. To further indicates that,  we conduct ablation analysis on different order in the supplementary materials, which further confirms this rationale.

\noindent\textbf{User Study.} 
We conducted a user study on 65 raters on 20 video pairs each to evaluate our method across three aspects: visual quality, motion quality, and text alignment. 
As shown in Table~\ref{tab:user_study}, the majority of participants rated our results as either superior ($\gg$, $>$) or comparable ($\approx$) to the baselines. 
Specifically, over 50\% of responses favored our approach in terms of visual quality, while motion quality and text alignment also received consistently positive feedback. The results confirm that our method achieves significant improvements in perceptual quality and semantic alignment, demonstrating its effectiveness in generating videos that are both visually appealing and faithful to textual descriptions.

\begin{table*}[t]
    \centering
    \scriptsize 
    \begin{minipage}[t]{0.35\linewidth}
        \centering
        \captionsetup{font=scriptsize, labelfont=bf}
        \caption{User study results across three aspects (Ours Vs Wan).}
        \label{tab:user_study}
        \renewcommand{\arraystretch}{0.9}
        \setlength{\tabcolsep}{4pt}
        \begin{tabular}{l|ccccc}
            \toprule
            Aspect & $\gg$ & $>$ & $\approx$ & $<$ & $\ll$ \\
            \midrule
            Visual Quality   & 21.5 & 30.3 & 31.3 & 12.7 & 4.2 \\
            Motion Quality   & 18.7 & 28.9 & 33.5 & 13.6 & 5.3 \\
            Text Alignment   & 23.2 & 29.8 & 30.1 & 11.9 & 5.0 \\
            \bottomrule
        \end{tabular}
    \end{minipage}
    \hfill
    \begin{minipage}[t]{0.5\linewidth}
        \centering
        \captionsetup{font=scriptsize, labelfont=bf}
        \caption{Ablation study on Wan2.1-T2V-1.3B on $\tau$ and $\beta$, results are reported on VideoAlign metrics VQ, MQ, and TA.}
        \label{tab:tau_beta_ablation}
        \renewcommand{\arraystretch}{0.9}
        \setlength{\tabcolsep}{3pt}
        \begin{tabular}{c|ccc|c|ccc}
            \toprule
            $\tau_{1,2}$ & VA & MQ & TA & $\beta$ & VA & MQ & TA \\
            \midrule
            0.60 & 3.49 & 0.30 & -0.70 & 0.1 & 3.48 & 0.36 & -0.73 \\
            0.75 & 3.50 & 0.31 & -0.71 & 0.3 & 3.49 & 0.31 & -0.70 \\
            0.80 & 3.53 & 0.29 & -0.79 & 0.5 & 3.51 & 0.28 & -0.68 \\
            \bottomrule
        \end{tabular}
    \end{minipage}
\end{table*}

\label{sec:discussion}

%% file: sections/6_conclusion.tex
\section{Conclusion}
In this work, we revisited the role of reward signals in GRPO-based video post-training and highlighted two recurring failure modes: shortcut-driven optimization under composite objectives and loss of discriminability along the training trajectory. Guided by these observations, we introduced TaRoS, a Target-Robust Reward Signaling framework that dynamically integrates multi-aspect reward components and adaptively suppresses saturated signals. This design mitigates reward hacking, postpones reward saturation, and provides stable optimization directions throughout training. Extensive experiments on Wan2.1-T2V and HunyuanVideo demonstrate that TaRoS consistently improves visual quality, motion coherence, and text-video alignment across diverse reward models. We will explore more sample-efficient post-training strategies in future work.

\label{sec:conclusion}